\documentclass[journal]{IEEEtran}

\ifCLASSINFOpdf
\else
   \usepackage[dvips]{graphicx}
\fi
\usepackage{url}
\usepackage{graphicx}
\usepackage{CJKutf8}
\usepackage{caption}
\usepackage{algorithm}
\usepackage{algorithmic}
\usepackage{multirow}
\usepackage{amsmath}
\usepackage{verbatimbox}
\usepackage{color}
\begin{document}

\title{Improving Mandarin End-to-End Speech Recognition with Word N-gram Language Model}

\author{Jinchuan Tian, Jianwei Yu, Chao Weng, Yuexian Zou, \IEEEmembership{Senior Member, IEEE} and Dong Yu, \IEEEmembership{Fellow, IEEE}
\thanks{Jinchuan Tian and Yuexian Zou are with the Advanced data and signal
processing laboratory, School of Electric and Computer Science, Peking
University, Shenzhen Graduate School, Shenzhen, China. This work was done when
Jinchuan Tian was an intern at Tencent AI Lab.}
\thanks{Jianwei Yu, Chao Weng and Dong Yu are with Tencent AI LAB. Jianwei Yu and Chao Weng are also from Tencent ASR Oteam.}
\thanks{Jianwei Yu is the corresponding author. (tomasyu@tencent.com)}
\thanks{Paper submitted to IEEE Signal Processing Letter (SPL)}
}

\markboth{Journal of \LaTeX\ Class Files, Vol. 14, No. 8, August 2015}
{Shell \MakeLowercase{\textit{et al.}}: Bare Demo of IEEEtran.cls for IEEE Journals}
\maketitle

\begin{abstract}
Despite the rapid progress of end-to-end (E2E) automatic speech recognition (ASR), it has been shown that incorporating external language models (LMs) into the decoding can further improve the recognition performance of E2E ASR systems. 
To align with the modeling units adopted in E2E ASR systems, subword-level (e.g., characters, BPE) LMs are usually used to cooperate with current E2E ASR systems. 
However, the use of subword-level LMs will ignore the word-level information, which may limit the strength of the external LMs in E2E ASR.
Although several methods have been proposed to incorporate word-level external LMs in E2E ASR, these methods are mainly designed for languages with clear word boundaries such as English and cannot be directly applied to languages like Mandarin, in which each character sequence can have multiple corresponding word sequences.
To this end, we propose a novel decoding algorithm where a word-level lattice is constructed on-the-fly to consider all possible word sequences for each partial hypothesis.
Then, the LM score of the hypothesis is obtained by intersecting the generated lattice with an external word N-gram LM.
The proposed method is examined on both Attention-based Encoder-Decoder (AED) and Neural Transducer (NT) frameworks. 
Experiments suggest that our method consistently outperforms subword-level LMs, including N-gram LM and neural network LM. 
We achieve state-of-the-art results on both Aishell-1 (CER 4.18\%) and Aishell-2 (CER 5.06\%) datasets and reduce CER by 14.8\% relatively on a 21K-hour Mandarin dataset.
\end{abstract}


\IEEEpeerreviewmaketitle

\section{Introduction}

\IEEEPARstart{E}{nd-to-End} automatic speech recognition (ASR) has made great progress in the past few years. Although E2E ASR frameworks such as attention-based encoder-decoder (AED)  \cite{las, lasctc} and neural transducer (NT)  \cite{rnnt} can directly map speech to token sequences through a single neural network, it has been found that incorporating an external language model into an E2E ASR system sometimes is crucial to utilize a large amount of text corpus. To this end, several approaches have been proposed to effectively integrate external LMs in E2E ASR systems  \cite{google_fusion, shallowfusion, coldfusion, localfusion}. One of the most widely used approaches is called \textit{shallow fusion} \cite{shallowfusion}.

In \textit{shallow fusion}, the log probabilities of each candidate hypothesis obtained from the E2E model and the external LM are interpolated during the decoding. 
To allow effective on-the-fly interpolation, the external LMs used in \textit{shallow fusion} are usually constructed using the same subword level modeling units (e.g., characters, BPE \cite{bpe}) with E2E ASR systems,  especially for languages without clear word boundary such as Mandarin and Japanese.
However, the use of subword level LMs will drop the word level information, which will compromise the strength of external LMs.
To overcome this issue, several approaches have been proposed to incorporate word-level external LMs during inference under the paradigm of \textit{shallow fusion}  \cite{lookaheadlm, lookaheadlm2} and achieved significant performance improvement. 
In \cite{lookaheadlm}, each hypothesis is firstly scored by the character-level LM until a word boundary is encountered. Then the known words are further rescored using the word-level LM while the character-level LM provides the scores for out-of-vocabulary tokens.
In \cite{lookaheadlm2}, a look-ahead mechanism is proposed to get rid of character-level LMs and score the hypothesis using a word-level RNN-LM only.
However, most of the previous methods require clear word boundaries to compute the word-level LM scores, which limits their application for those languages with blurred word boundaries, such as Mandarin and Japanese. 
Therefore, we argue that the way to effectively incorporate word-level LMs into an E2E ASR system for languages without clear word boundaries needs further investigation.

To utilize the word-level LM during decoding, a subword-to-word conversion is a prerequisite. However, a character-level sequence can be mapped to different word sequences.
Therefore, the main issue that needs to be addressed in incorporating word-level LM in E2E ASR for languages like Mandarin is: how to consider all candidate word sequences of a partial character hypothesis. 
In this work, we propose a new method to integrate word N-gram LMs into the decoding of E2E ASR systems without the requirement of clear word boundaries.
During decoding, each character-level partial hypothesis generated by the ASR system is firstly converted into a word lattice. Each path of the lattice stands for one possible subword-to-word conversion result. 
Then the score of each path is obtained by \textit{intersecting} the word lattice with an external weighted finite-state acceptor (WFSA) word-level N-gram LM.
Finally, the best score among all the paths (or the averaged score of all the paths) in the lattice is used as the score computed with word-level LM for the given partial hypothesis. 
In this way, the proposed method is free from word boundary and considers all possible subword-to-word conversion results.

We examine the proposed method on two of the most popular E2E ASR frameworks: AED and NT. Exhaustive experiments show that our method consistently outperforms its subword-level counterparts, including both the N-gram LM and neural network LM. Superior robustness to hyper-parameters is also observed in our method. Experimentally, we achieve the character error rate (CER) of 4.18\% and 5.06\% on Aishell-1 \cite{aishell1} \textit{test} set and Aishell-2 \cite{aishell2} \textit{test-ios} set. Our method is also compatible with large-scale ASR tasks: up to 2\% absolute, or 14.8\% relative, CER reduction is observed on a 21k-hour Mandarin corpus.

The main contributions of this work are concluded below: 
1) we propose a novel method to incorporate external word-level N-gram language models into E2E ASR systems for languages without clear word boundaries;
2) To the best of our knowledge, the proposed method achieves state-of-the-art results on two of the widely used Mandarin datasets: Aishell-1 and Aishell-2.
\vspace{-5pt}
\section{On-the-fly Decoding with Word N-gram}

\subsection{Shallow fusion in AED and NT}
\textit{Shallow fusion} \cite{shallowfusion} is a common method for LM adaptation of E2E ASR systems. In this work, we still follow the paradigm of \textit{shallow fusion} to integrate the word-level N-gram LM. Without loss of generality, we assume a character-level partial hypothesis $\mathbf{C}_1^m$ is proposed by the E2E ASR system, where $\mathbf{C}_1^{m-1}$ is the history context and $c_m$ is the newly proposed character. 
In \textit{shallow fusion}, the LM log-posterior $\log p(c_m|\mathbf{C}_1^{m-1})$ is required during decoding for various E2E ASR systems. Specifically, for the decoding algorithm\cite{lasctc} of AED, assume $\delta(\mathbf{C}_1^m)$ is the score of partial hypothesis $\mathbf{C}_1^m$. In \textit{shallow fusion}, the score is computed recursively:
\begin{equation}
\setlength\abovedisplayskip{0.2cm}
\setlength\belowdisplayskip{0.2cm}
    \delta(\mathbf{C}_1^m) = \delta(\mathbf{C}_1^{m-1}) + [s^{\tt{AED}}_{c_m} + \beta_{\tt{AED}} \cdot \log p(c_m|\mathbf{C}_1^{m-1})]
\end{equation}
Likewise, for the decoding algorithm ALSD\cite{alsd} of NT system, assume $\delta_t(\mathbf{C}_1^m)$ is the score of partial hypothesis $\mathbf{C}_1^m$ with time stamp $t$ and the LM log-posterior is adopted only when a non-blank character is generated:
\begin{equation}
\setlength\abovedisplayskip{0.2cm}
\setlength\belowdisplayskip{0.2cm}
    \delta_t(\mathbf{C}_1^m) = \delta_t(\mathbf{C}_1^{m-1}) + [s^{\tt{NT}}_{c_m, t} + \beta_{\tt{NT}} \cdot \log p(c_m|\mathbf{C}_1^{m-1})]
\end{equation}
$s^{\tt{AED}}_{c_m}$ and $s^{\tt{NT}}_{c_m, t}$ are the scores for character $c_m$ generated by AED and NT systems respectively while $\beta_{\tt{AED}}$ and $\beta_{\tt{NT}}$ are the hyper-parameters in decoding.
\vspace{-8pt}
\subsection{Word-level N-gram LM integration}
\label{score}
For character-level LMs, the log-posterior $\log p(c_m|\mathbf{C}_1^{m-1})$ used in \textit{shallow fusion} is always provided. For word-level LM, however, this log-posterior cannot be accessed directly. Instead, we compute the character-level log-posterior as follow:
\begin{equation}
    \label{autoregressive}
    \setlength\abovedisplayskip{0cm}
    \setlength\belowdisplayskip{0cm}
    \log p(c_m|\mathbf{C}_1^{m-1}) = \log P(\mathbf{C}_1^m) - \log P(\mathbf{C}_1^{m-1})
\end{equation}
As suggested in the equation, the word-level LM can be integrated into \textit{shallow fusion} as long as the sequential probability $P(\mathbf{C}_1^m)$ of any character sequence $\mathbf{C}_1^m$ can be computed by a word-level LM. Below we explain how $P(\mathbf{C}_1^m)$ is obtained by a word-level N-gram LM.

To assess character sequence by a word-level LM, a character-to-word conversion is firstly required. For languages with explicit word boundary (a.k.a., \textit{blank}) such as English, the word sequence converted from the corresponding character sequence is unique. However, for other languages like Mandarin, the absence of clear word boundary means a character sequence can be converted into multiple different word sequences. E.g., the Mandarin character sequence {\begin{CJK}{UTF8}{gbsn}孙悟空 can be mapped to word sequences $[$孙,悟,空$]$, $[$孙,悟空$]$ and $[$孙悟空$]$. Note we assume word 孙悟 is not a valid Mandarin word so the sequence $[$孙悟,空$]$ is invalid and not considered.\end{CJK}}

Formally, note $f$ as a function that converts character-level sequence $\mathbf{C}_1^{m}$ into a word-level sequence $\mathbf{W}_1^{n_f}$:
\begin{equation}
    \setlength\abovedisplayskip{0cm}
\setlength\belowdisplayskip{0cm}
    \mathbf{W}_1^{n_f} = f(\mathbf{C}_1^{m}) = [w_1, ..., w_{n_f}] \quad \forall\; w_i \in \mathcal{V}
\end{equation}
where $\mathcal{V}$ is a known word vocabulary and $n_f$ is the length of the word-level sequence that depends on $f$. Thus, the probability of the character-level sequence $P(\mathbf{C}_1^m)$ is formulated as the weighted sum probability of all possible word-level sequence $\mathbf{W}_1^{n_f}$:
\begin{equation}
\label{eq_subwd_p}
    \setlength\abovedisplayskip{0cm}
    \setlength\belowdisplayskip{0cm}
    P(\mathbf{C}_1^{m}) = \sum_{f\in \mathcal{F}} P(f)P(f(\mathbf{C}_1^{m})) = \sum_{f\in \mathcal{F}} P(f)P(\mathbf{W}_1^{n_f})
\end{equation}
where $\sum_{f} P(f) = 1$ and $\mathcal{F}$ is the set of all valid character-to-word conversion functions. 

\begin{figure}
    \centering
    \includegraphics[width=6cm]{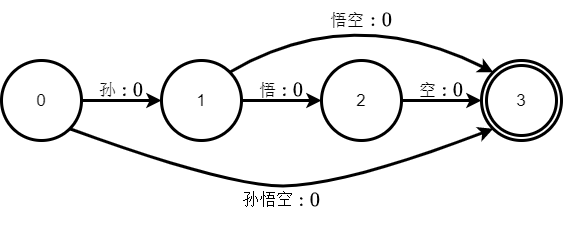}
    \caption{\begin{CJK}{UTF8}{gbsn}Lattice built from character-level sequence 孙悟空 and vocabulary $\mathcal{V}=$ \{孙, 悟，空, 悟空, 孙悟空\}. All input symbols are in word-level.\end{CJK}} 
    \vspace{-15pt}
    \label{fig_lattice}
\end{figure}

To fully explore all possible word sequences $\mathbf{W}_1^{n_f}$ and their corresponding character-to-word conversion functions $f$ in Eq. \ref{eq_subwd_p}, a lattice-based method is proposed. The character sequence $\mathbf{C}_1^{m}$ is firstly transformed into a lattice $\mathcal{Q}$ (which is also a WFSA). For any word sequence $\mathbf{W}_1^{n_f}$, there is one and only one path in the lattice whose input label sequence is exactly $\mathbf{W}_1^{n_f}$. Thus, the whole lattice is a representation of all possible character-to-word conversion results. An example lattice of character sequence {\begin{CJK}{UTF8}{gbsn}孙悟空\end{CJK}} is shown in Fig.\ref{fig_lattice}.

Given the character sequence $\mathbf{C}_1^{m}$ and word vocabulary $\mathcal{V}$, the detailed building process of the lattice is in Algorithm 1.
For any continuous $r$ characters $\mathbf{C_{s}^{s+r-1}}$, if they can form a word in $\mathcal{V}$ (a.k.a, $w=\mathbf{C_{s}^{s+r-1}}\in \mathcal{V}$), an arc from state $s-1$ to state $s+r-1$ is added to the arc set $\mathcal{A}$. 
The weights on all these arcs are 0. 
Once all valid $s$ and $r$ are explored, the lattice is built from the arc set $\mathcal{A}$. 
In addition, we ensure that any character predicted by the ASR systems is also in the word vocabulary $\mathcal{V}$ ($\forall \; c_i \in \mathcal{V}$) so that all the out-of-vocabulary (OOV) words can always be considered as the sequence of single characters and the end state (state $m-1$) is always reachable. 


\begin{algorithm} 
	\caption{Build lattice for character-level sequence $\mathbf{C}_1^{m}$} 
	\label{algo_lattice} 
	\begin{algorithmic}
		\REQUIRE character-level sequence $\mathbf{C}_1^{m}$, word-level vocabulary $\mathcal{V}$, maximum word length $l$.
		\STATE arc set $\mathcal{A} \gets \emptyset$
		\FOR {$r = 1, .., l$}
		\FOR {$s = 1, ..., m - r$}
		\IF {$\mathbf{C_{s}^{s+r-1}} \in \mathcal{V}$}
		\STATE $\mathcal{A} \gets \mathcal{A}\ \cup$ arc$(s-1, s+r-1, \mathbf{C_{s}^{s+r-1}}, 0)$
		\ENDIF
		\ENDFOR
		\ENDFOR
		\RETURN WFSA($\mathcal{A}$)
	\end{algorithmic} 
\end{algorithm}

Before decoding, the word-level N-gram LM is trained following \cite{manninglm, stolcke2002srilm} and is transformed into a WFSA (termed as $\mathcal{G}$). In the decoding stage, once the lattice $\mathcal{Q}$ for character sequence $\mathbf{C}_1^{m}$ is built, the character-level probability in eq.\ref{eq_subwd_p} is computed through an FST-based method. A new WFSA $\mathcal{QG}$ is generated by the word lattice $\mathcal{Q}$ and the word N-gram $\mathcal{G}$:
\begin{equation}
    \setlength\abovedisplayskip{0cm}
    \setlength\belowdisplayskip{0cm}
    \mathcal{QG} = \mathcal{Q} \circ \mathcal{G}
\end{equation}
where $\circ$ denotes \textit{intersect} operation\cite{intersect} in WFSA. Before \textit{intersect}, we add self-loops with 0 weight and \textit{epsilon} input label to explore all backoff paths in $\mathcal{G}$. 
Subsequently, the character-level sequence probability $P(\mathbf{C}_1^{m})$ is equivalent to the forward score of the end state in $\mathcal{QG}$. 
\begin{equation}
    \log  P(\mathbf{C}_1^{m})  = \log \sum_{f\in \mathcal{F}} P(f)P(\mathbf{W}_1^{n_f}) = \text{Forward\_Score}(\mathcal{QG})
\end{equation}

In FST literature\cite{horifst}, \textit{semirings} are the algebraic structures that define the rule of forward score computation. In this work, we consider two well-known \textit{semirings}: \textit{tropical-semiring} and \textit{Log-semiring}.
\textit{Tropical-semiring} only considers the path with the best accumulated score, which is equivalent to set:
\begin{equation}
    \label{tropical}
  P(f) = \left\{
          \begin{aligned}
              & 1  \quad if \ f = \arg\mathop{\max}_{f'\in \mathcal{F}}P(f'(\mathbf{C}_1^{m}))\\
              & 0  \quad otherwise \\
          \end{aligned}
    \right\}
\end{equation} 
On the other hand, \textit{Log-semiring} equally accumulates the score of every path that can be accepted by $\mathcal{QG}$, which is equivalent to set: 
\begin{equation}
  \setlength\abovedisplayskip{0cm}
  \setlength\belowdisplayskip{0cm}
    \label{log}
  P(f) = \frac{1}{|\mathcal{F}|} 
\end{equation}

\section{Experiments}
In this section, we firstly describe the experimental setup. Then the experimental results and analysis on three Mandarin datasets are presented.
\vspace{-15pt}
\subsection{Experimental setup}
\label{expsetup}
1) \textbf{Datasets}: We examine our method on two popular open-source Mandarin datasets: Aishell-1 (178 hours), Aishell-2 (1k hours). To further evaluate the effectiveness of the proposed method on large-scale ASR tasks, we also apply our method to a 21k-hour Mandarin dataset and report its performance under 6 test scenarios: reading, translation, guild, television, music, education. \\
2) \textbf{E2E ASR baselines}: We verify the effectiveness of our method on two of the most popular E2E ASR systems: Attention-based Encoder-Decoder (AED) and Neural Transducer (NT). For AED, a Conformer\cite{conformer} encoder and a Transformer\cite{transformer} decoder are adopted, which consumes 46M parameters; for NT, a Conformer encoder, an LSTM prediction network and an MLP joint network are used (89M parameters). Following our previous work \cite{tian2021consistent}, the LF-MMI criterion\cite{lfmmi_16,lfmmi_18} is adopted in the training stage of the two systems for better performance. In the decoding stage, we adopt the MMI Prefix Score and MMI Alignment Score algorithms for AED and NT respectively \cite{tian2021consistent}. The beam size of decoding is fixed to 10 and the decoding algorithms used in AED and NT systems are described in  \cite{lasctc} and  \cite{alsd}. The E2E ASR baselines are mainly implemented by Espnet \cite{espnet}. \\ 
3) \textbf{Language Models}: We compare the performance of three types of LMs: word-level N-gram LM, character-level N-gram LM and character-level neural network LM (NNLM). 
All LMs used in Aishell-1 and Aishell-2 are trained from transcriptions of the training set and adopt no external resources. An additional text corpus is used in the 21k-hour experiments.
Before training the word-level LM, the transcriptions are segmented into word-level using jieba\footnote{https://github.com/fxsjy/jieba}. All N-gram LMs are trained by kaldi\_lm\footnote{https://github.com/danpovey/kaldi\_lm} while all NNLMs use a standard 1-layer LSTM with the hidden size of 650.
Unless otherwise specified, the default settings for the decoding are described as follows. The LM weight of  \textit{shallow fusion} method (a.k.a, $\beta_{\tt{AED}}$ and $\beta_{\tt{NT}}$) is fixed to 0.4 when using word-level N-gram LM. However, we enumerate the LM weights of character-level N-gram LMs and NNLMs in set \{0.2, 0.4, 0.6, 0.8, 1.0, 1.2, 1.5, 2.0\} and report the best results. Also, the default order of word-level N-gram LM is 3 and the \textit{log-semiring} is used. The WFSA-related operations are implemented by K2\footnote{https://github.com/k2-fsa/k2}. \\
\vspace{-20pt}
\subsection{Results on Aishell-1 and Aishell-2}
\label{res_aishell}

\begin{table*}[htpb]
    \centering
    \scalebox{1.0}{
    \begin{tabular}{|c|c|c|c|c|c|c|c|c|c|c|c|}
    \hline
         \multirow{2}{*}{No.} &\multirow{2}{*}{System} & \multirow{2}{*}{LM type}           & \multirow{2}{*}{N-gram order} & \multirow{2}{*}{Semiring}  &  \multirow{2}{*}{LM weight}  &  \multicolumn{2}{|c|}{Aishell-1}  & \multicolumn{3}{|c|}{Aishell-2} & RTF(CPU)\footnotemark[4] \\
         \cline{7-12}
                              &                        &                                    &                               &                                     &                              &  dev & test                       & ios & android & mic  & Aishell2-ios \\
         \hline
         1                    & \multirow{11}{*}{AED}   & -                                  & -                             & -                                   &  0.0                         & 4.55 & 5.10                       & 5.93 & 7.04 & 6.79 &               \bf{1.45} \\
         \cline{3-12}
         2                    &                        & \multirow{7}{*}{Word N-gram}       & \multirow{5}{*}{3}            & \multirow{5}{*}{log (Eq.\ref{log})} &  0.2                         & 4.22 & 4.62                       & 5.43 & 6.48 & 6.21 &                2.03 \\
         3                    &                        &                                    &                               &                                     &  0.4                         & \bf{4.08} & 4.45                  & 5.26 & \bf{6.22} & \bf{5.92} & 2.02 \\
         4                    &                        &                                    &                               &                                     &  0.6                         & 4.09 & \bf{4.43}                  & 5.27 & \bf{6.22} & 5.99 &           2.00 \\ 
         5                    &                        &                                    &                               &                                     &  0.8                         & 4.20 & 4.58                       & 5.56 & 6.53 & 6.26 &                2.03 \\
         6                    &                        &                                    &                               &                                     &  1.0                         & 4.42 & 4.80                       & 6.00 & 6.94 & 6.77 &                2.04 \\
         \cline{4-12}
         7                    &                        &                                    & 3                             & tropical (Eq.\ref{tropical})        &  0.4                         & \bf{4.08}  & 4.45                 & 5.26 & \bf{6.22} & \bf{5.92} & 2.01\\
         8                    &                        &                                    & 4                             & log (Eq.\ref{log})                  &  0.4                         & \bf{4.08}  & 4.46                 & 5.26 & \bf{6.22} & 5.93 &      2.08\\
         \cline{3-12}
         9                   &                         & Word N-gram Rescore                & 3                             & log (Eq.\ref{log})                  &  (0, 2.0]                    & 4.11 & 4.47                       & 5.31 & 6.27 & 6.03 &                1.45 \\
         10                    &                       & Char. N-gram                       & 6                             & -                                   &  (0, 2.0]                    & 4.24 & 4.63                       & \bf{5.25} & 6.17 & 5.94 &                1.60 \\
         11                   &                        & Char. NN                           & -                             & -                                   &  (0, 2.0]                    & 4.41 & 4.64                       & 5.79 & 6.77 & 6.51 &                1.87 \\
         
         \hline\hline
         12                    & \multirow{11}{*}{NT}   & -                                  & -                             & -                                  &  0.0                        & 4.20 & 4.60                        & 5.41 & 6.56 & 6.39 &                \bf{0.50}\\
         \cline{3-12}
         13                    &                        & \multirow{7}{*}{Word N-gram}       & \multirow{5}{*}{3}            & \multirow{5}{*}{log (Eq.\ref{log})}&  0.2                        & 3.97 & 4.29                        & 5.13 & 6.15 & 6.05 &                4.12\\
         14                    &                        &                                    &                               &                                    &  0.4                        & \bf{3.87} & \bf{4.18}              & \bf{5.06} & \bf{6.08} & \bf{5.98} & 4.48\\
         15                    &                        &                                    &                               &                                    &  0.6                        & \bf{3.87} & 4.22                   & 5.16 & 6.23 & 6.09 &                4.43\\ 
         16                    &                        &                                    &                               &                                    &  0.8                        & 4.00 & 4.33                        & 5.37 & 6.36 & 6.23 &                4.25\\
         17                    &                        &                                    &                               &                                    &  1.0                        & 4.25 & 4.55                        & 5.65 & 6.71 & 6.55 &                4.26\\
         \cline{4-12}
         18                    &                        &                                    & 3                             & tropical (Eq.\ref{tropical})       &  0.4                         & \bf{3.87} & \bf{4.18}             & \bf{5.06} & \bf{6.08} & \bf{5.98} & 4.59\\
         19                    &                        &                                    & 4                             & log (Eq.\ref{log})                 &  0.4                         & \bf{3.87} & \bf{4.18}             & \bf{5.06} & \bf{6.08} & 6.00 &      4.50\\
         \cline{3-12}
         20                    &                        & Word N-gram Rescore                & 3                             & log (Eq.\ref{log})                 &  (0, 2.0]                   & 3.93 & 4.27                         & 5.11 & \bf{6.08} & 6.02 & 0.50\\
         21                    &                        & Char. N-gram                       & 6                             & -                                  &  (0, 2.0]                   & 4.11 & 4.23                         & 5.28 & 6.36 & 6.27 &      0.90\\
         22                    &                        & Char. NN                           & -                             & -                                  &  (0, 2.0]                   & 4.09 & 4.43                         & 5.33 & 6.41 & 6.28 & 1.62\\
         
         \hline
    \end{tabular}
    }
    \caption{Results on Aishell-1 and Aishell-2 datasets. CER and RTF are reported. All LMs are adopted using \textit{shallow fusion}.}
    \label{tab_main}
    \vspace{-15pt}
\end{table*}

We present our experimental results on Aishell-1 and Aishell-2 in table \ref{tab_main}. Our main observations are listed below. \\
1) \textbf{LM weight of shallow fusion}: The LM weight of \textit{shallow fusion} is the first hyper-parameter we investigate. As shown in exp.1-6 and exp.12-17, LM weight of 0.4 consistently shows the best improvement on both ASR systems, which explains our default choice of this weight. Also, our method is robust to the choice of LM weight: compared with the baseline systems (exp.1, exp.12), CER reduction is consistently observed when the LM weight is smaller or equal to 0.8. \\
2) \textbf{Semiring}: For both E2E ASR frameworks, the performance difference caused by the choice of the semiring is negligible (exp.3 vs. exp.7; exp.14 vs. exp.18). \\
3) \textbf{Order of word-level N-gram LM}: We find word-level N-gram LMs of higher order do not show further performance improvement (exp.3 vs. exp.8; exp.14 vs. exp.19). 
One explanation of this is that: since we only use the training transcriptions of Aishell datasets, the higher-order terms (e.g., 4-gram terms) seen in LM training are rarely seen in the test set, as the training transcriptions are small in volume and achieve poor coverage. \\ 
4) \textbf{Comparison with Word N-gram Rescore}: The sequence-level LM scores computed by Eq.\ref{eq_subwd_p} can also be used for rescoring after the N-best hypotheses are proposed by ASR systems. However, slight performance degradation is observed if the word N-gram LM is adopted in rescoring rather than in on-the-fly decoding (exp.3 vs. exp.9; exp.14 vs. exp.20). \\
5) \textbf{Comparison with character-level LM}: We compare the performance of our method with those character-level LMs, including both the N-gram LM and the neural network LM (exp.3 vs. exp.10-11; exp.14 vs. exp.21-22). Although the character-level LMs do provide some improvement on the recognition performance, our word-level N-gram LM outperforms the character-level counterparts in most test sets and frameworks. This observation validates our claim that the word-level information is beneficial during decoding. \\ 
6) \textbf{Different E2E ASR frameworks}: The results suggest that the proposed method is compatible with both AED and NT systems and achieves performance improvement consistently (exp.1,3 vs. exp.12,14). \\
7) \textbf{Real-time factor (RTF)}: Besides the recognition accuracy (CER\%), the real-time factor (RTF) is also concerned. We find that hyper-parameters like LM weight, LM order and semiring have negligible impact on the computational overhead. 
For AED, the computational overhead is close to that of NNLM (2.02 to 1.87). For NT, the RTF of our method is 4.48. The computational overhead on the NT system is much higher than that of the AED system since the NT systems assess the partial hypotheses for each frame and each word sequence length. The acceleration of our method on NT is left for future work.

\begin{table}[htpb]
    \centering
    \begin{tabular}{|l|c|c|}
        \hline
        \multirow{2}{*}{System} & Aishell-1 & Aishell-2 \\
                              & dev/test & ios/android/mic \\
        \hline\hline
        SpeechBrain \cite{speechbrain}           &  -  / 5.58 & -    / -   / -        \\                                    
        Espnet \cite{espnet_result}              & 4.4 / 4.7  & 6.8  / 7.6 / 7.4      \\
        Wenet \cite{yao2021wenet}                &  -  / 4.36 & 5.35 / -   / -        \\
        Icefall                                 &  -  / 4.26 & -    / -   / -        \\
        \hline\hline  
        \multicolumn{3}{|l|}{Ours} \\
        \hline
        AED                              & 4.55      / 5.10             & 5.93      / 7.04      / 6.79      \\
        \ \  + Word N-gram                      & \bf{4.08} / \bf{4.45}   & \bf{5.26} / \bf{6.22} / \bf{5.92} \\
        \hline
        NT                               & 4.20      / 4.60                   & 5.41        / 6.56      / 6.39 \\
        \ \ + Word N-gram                      & \bf{3.86} / \bf{4.18}    & \bf{5.06} / \bf{6.08} / \bf{5.98} \\
        \hline
    \end{tabular}
    \caption{Results on Aishell-1 and Aishell-2 datasets. All results are in CER\%. No external resources are used.}
    \label{tab_global}
\end{table}
\footnotetext[4]{single thread; Pytorch implementation; Intel Xeon 8255C CPU, 2.5GHz} 
\vspace{-5pt}
We finally conclude our experimental results and compare them with other competitive frameworks in table \ref{tab_global}. As suggested in the table, significant CER reduction is consistently observed in both E2E ASR frameworks and various test sets. The maximum and minimum absolute CER reduction are 0.87\% (AED, Aishell-2 mic) and 0.34\% (NT, Aishell-1 dev) respectively. The NT model with our word-level N-gram LM achieves all of the best results except that on the Aishell-2 mic set. Our best results on Aishell-1 test set and Aishell-2 test-ios set are 4.18\% and 5.06\% respectively. To the best of our knowledge, these are the state-of-the-art results on these two datasets. 
The success of our method emphasizes that the adoption of word-level N-gram LM is promising. 

\footnotetext[5]{This research is partially supported by NSFC (No:6217021843) and Shenzhen Science \& Technology Fundamental Research Programs (No:JCYJ20180507182908274 \& JSGG20191129105421211).}

\vspace{-10pt}
\subsection{Results on 21k-hour Mandarin dataset}
\label{res_oteam}
We show that the proposed method is still effective for large-scale ASR tasks and external text corpus. Specifically, we train an NT model on the 21k-hour speech data and the word-level LM (3-gram, with the size of 3.4G) is constructed from an external text corpus. 
As shown in table \ref{tab_oteam}, incorporating word-level N-gram LM based on the proposed method consistently achieves better recognition performance on all 6 test sets. 
The absolute CER reduction is 1.1\% on average while the maximum relative CER reduction is 14.8\% (in \textit{reading} set). 

\begin{table}[htb]
    \centering
    \begin{tabular}{l|cccccc|c}
          \hline
          \hline 
          Test set      & re          & tr          & gu          & tv          & mu          & ed          & Mean    \\
          \hline
          NT                    & 5.4         & 17.8        & 19.0        & 9.2         & 14.5        & 11.8        & 13.0       \\
          \ \ + Word N-gram     & \bf{4.6}    & \bf{16.4}   & \bf{17.0}   & \bf{8.4}    & \bf{13.3}   & \bf{11.4}   & \bf{11.9}     \\
          \hline
           
          \hline
    \end{tabular}
    \caption{Experimental results on a 21k-hour Mandarin with w/o external word-level N-gram LM. re: reading; tr: translation; gu: guild; tv: television; mu: music; ed: education.}
    \label{tab_oteam}
\end{table}

\vspace{-15pt}
\section{Conclusion}
In this work, we propose a new method to integrate the word-level N-gram language models into the decoding process of end-to-end automatic speech recognition systems. Compared with the language models that work in subword-level, our method consistently achieves better performance. Also, our method is robust to various hyper-parameters and is applicable to large-scale ASR tasks. Experimentally, we achieve state-of-the-art results on two of the most popular Mandarin datasets.


\bibliographystyle{IEEEtran}
\bibliography{./refs.bib}

\end{document}